\documentclass[letterpaper]{article} 
\usepackage[preprint]{aaai2027} 
\usepackage[hyphens]{url} 
\usepackage{graphicx} 
\usepackage{natbib} 
\usepackage{caption} 
\usepackage{booktabs}
\usepackage{amsmath}
\usepackage{amssymb}
\usepackage{pifont}
\usepackage{soul}
\usepackage{float}

\newif\ifshowcomments
\showcommentstrue
\soulregister\cite7

\title{How Robust Are LLMs to Vietnamese Dialects?}
\author{
    Minh Tran\textsuperscript{\rm 1}\thanks{Equal contribution.}\corresponding,
    Trinh Chau\textsuperscript{\rm 2}\footnotemark[1],
    Thanh-Nhan Le\textsuperscript{\rm 3},
    Nam Tran\textsuperscript{\rm 4},\\
    Luan Thanh Nguyen\textsuperscript{\rm 5},
    Cuong Dang\textsuperscript{\rm 6},
    Duc Hoang\textsuperscript{\rm 7}
}
\affiliations{
    \textsuperscript{\rm 1}University of Science, VNU-HCM,
    \textsuperscript{\rm 2}VNU University of Engineering and Technology\\
    \textsuperscript{\rm 3}New York University Abu Dhabi, \textsuperscript{\rm 4}Independent Researcher\\
    \textsuperscript{\rm 5}University of Information Technology, VNU-HCM\\
    \textsuperscript{\rm 6}Virginia Tech, \textsuperscript{\rm 7}MIT
}
\begin{document}
\maketitle

\begin{abstract}
Large Language Models (LLMs) are typically evaluated on standard written Vietnamese, yet everyday communication frequently involves regional dialects that preserve meaning but differ in surface form. Existing Vietnamese dialect work largely addresses this issue through dialect-to-standard normalization instead of measuring how the model fails under Vietnamese dialectal inputs. To address this gap, we present the first study that systematically evaluates the robustness of LLMs to Vietnamese dialect variation across multiple tasks, quantifying dialect-induced performance degradation and analyzing failure patterns behind it. We introduce \textbf{VialectBench} (Vietnamese Dialects Benchmarking), a controlled benchmark for testing whether model decisions remain stable when task content is expressed in six Vietnamese dialect groups. \textbf{VialectBench} contains 400 Standard Vietnamese source instances and 2,400 human-written dialectal rewrites spanning emotion recognition (ER), natural language inference (NLI), question answering (QA), and multiple-choice question answering (MCQA). Dataset evaluation with a fixed reference language model shows that the dialectal rewrites induce a measurable model-relative likelihood shift while remaining nearly equal in length to their Standard counterparts. Across ten instruction-tuned models, dialectal inputs reduce average performance by $2.82\%$, and no evaluated model is fully dialect-invariant. All four tasks are affected, with QA showing the largest average degradation. Robustness also varies substantially across dialect groups: PNT3 and PNT2 cause the largest average performance drops, at $6.17\%$ and $4.73\%$, respectively, whereas PNB slightly improves average performance by $0.42\%$.The Central dialect group (PNT1--PNT4) also yields the highest average harmful-flip rate across all models, at $6.54\%$. These findings show that strong performance on Standard Vietnamese does not guarantee reliable behavior under meaning-preserving regional variation.
\end{abstract}
\section{Introduction}

\begin{figure}[!t]
    \centering
    \includegraphics[width=0.85\linewidth]{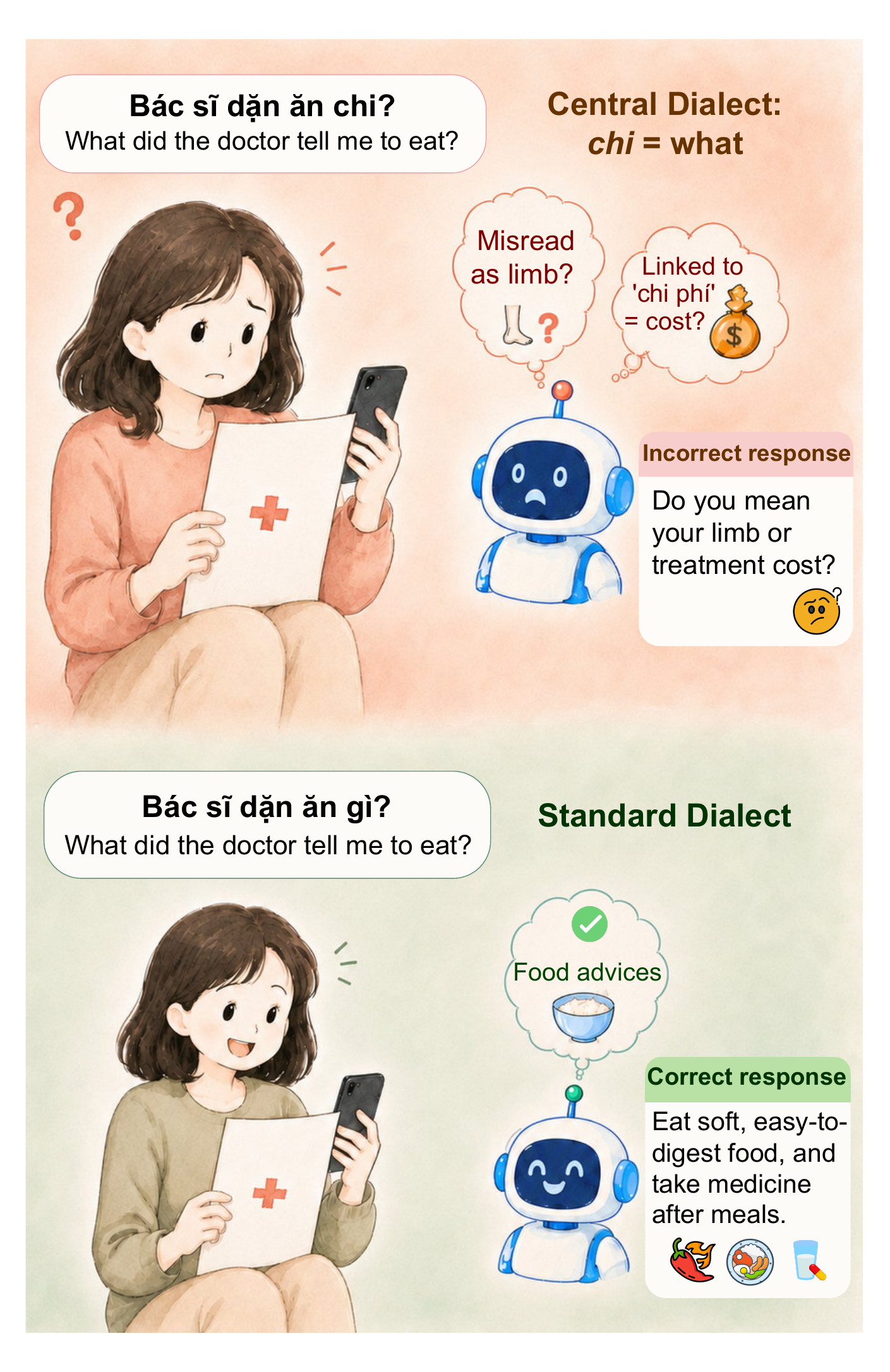}
    \caption{Overview of how Vietnamese dialectal variation affects LLM interpretation and leads to unfaithful responses.}
    \label{fig:overview}
\end{figure}

Large Language Models (LLMs) have become a dominant paradigm in natural language processing (NLP), achieving state-of-the-art performance across a wide range of tasks~\cite{zhao2026surveylargelanguagemodels}. However, their reliability remains fragile under natural input variation, including changes in wording, style, and linguistic form, which can lead to performance degradation~\cite{zhu2024promptrobustevaluatingrobustnesslarge}. Among these forms of natural variation, dialectal variation is especially important. Unlike synthetic perturbations, dialectal variation reflects language use by real communities, yet it can still shift model predictions and amplify fairness or safety concerns~\cite{hofmann2024aigenerates,faisal2024dialectbench}. Recent benchmarks have studied this problem in English, Arabic, and multilingual settings~\cite{ziems2022multivalue,mousi-etal-2025-aradice,faisal2024dialectbench,lin2025redial,gupta2025endive}, consistently showing that models performing well on standardized language may remain unreliable when processing regional varieties.

This issue is especially important for Vietnamese, where substantial regional variation exists across Northern, Central, and Southern dialects, with further diversity at the provincial level. These varieties differ in lexical choices, pronouns, discourse particles, interrogatives, and function words, many of which can be task-critical for language understanding. Figure~\ref{fig:overview} illustrates how a meaning-preserving dialectal expression can still lead to model misunderstanding.

\noindent\textbf{Research Gaps and Significance.} \ding{182} Current Vietnamese LLM benchmarks predominantly evaluate models using standardized written Vietnamese, which is closely associated with the Northern variety~\cite{bui-etal-2025-vmlu,ho2020emotionrecognitionvietnamesesocial,vanhuynh2025newbenchmarkdatasetmixtureofexperts,Luu_2025,Nguyen_2022}, rather than controlled regional dialectal variants. Consequently, it remains unclear whether LLMs produce consistent predictions when prompted with meaning-equivalent regional Vietnamese forms rather than standard Vietnamese. \ding{183} Existing Vietnamese dialect studies have mainly framed regional variation as a dialect-to-standard normalization problem~\cite{le2023centralnorthern,ta2026vidia2std}. \citet{le2023centralnorthern} introduce a Central--Northern parallel corpus, while ViDia2Std~\cite{ta2026vidia2std} provides a nationwide corpus covering all 63 provinces. These resources show that normalizing dialectal inputs can benefit downstream tasks such as sentiment analysis and machine translation. However, they do not directly test whether LLMs themselves are robust to meaning-preserving dialectal inputs. Because they do not provide multiple dialectal variants of the same gold-labeled source instance, these datasets lack the controlled structure required for comparable model decisions across dialects and therefore cannot quantify dialect-induced performance drops, prediction flips, or task-specific failure patterns.

\textit{Our work fills this gap by introducing a unified robustness benchmark for evaluating LLMs across emotion recognition (ER), natural language inference (NLI), question answering (QA), and multiple-choice question answering (MCQA). Each standard Vietnamese instance is rewritten into six meaning-preserving regional dialectal variants, while preserving the meaning. Following this design, the standard and dialectal inputs differ primarily in regional linguistic form, allowing observed performance changes to be attributed to dialectal variation. Moreover, the selected tasks require different model capabilities, including reasoning, evidence extraction, and answer selection, allowing us to examine whether dialectal variation affects these abilities differently.}

\noindent\textbf{Novelties and Contributions.}
Our main contributions are summarized as follows:
\begin{enumerate}
    \item We introduce \textbf{VialectBench}, the first end-to-end human-annotated parallel benchmark for evaluating Vietnamese LLM robustness under meaning-preserving dialectal variation, covering six regional dialect groups and four language-understanding tasks.
    
    \item We introduce a controlled framework for evaluating performance and prediction flips across matched Standard Vietnamese and dialectal inputs in four tasks, and benchmark ten open- and closed-source instruction-tuned models across tasks and regions.
    
    \item We illustrate that no evaluated model is fully robust to Vietnamese dialect variation: all four task families are affected, QA exhibits the broadest degradation, and the Central subgroups PNT2 and PNT3 consistently produce the largest performance losses and harmful flips. These results demonstrate that strong Standard Vietnamese performance does not guarantee reliable behavior across regional varieties.
\end{enumerate}
\section{Related Work}
\noindent\textbf{Dialect Robustness of LLMs.} Recent work has begun to benchmark LLM performance under dialectal variation across multiple languages and tasks.

\emph{(i) English Dialect Benchmarks.} \citet{ziems2022multivalue} introduce Multi-VALUE, a framework for cross-dialectal English NLP using transformations grounded in linguistic features across 50 English varieties. \citet{srirag-etal-2025-evaluating} introduce M-MD3 to evaluate dialect robustness in conversational understanding, combining naturally occurring US and Indian English dialogues with dialect-transformed and normalized counterparts; their results reveal consistent performance gaps against Indian English in target-word prediction and selection. \citet{lin2025redial} present ReDial, which evaluates fairness and robustness on reasoning tasks using over 1,200 parallel Standard English--AAVE query pairs from benchmarks such as HumanEval and GSM8K, demonstrating broad LLM brittleness on AAVE reasoning inputs. \citet{gupta2025endive} extend this to five underrepresented English dialects across 12 datasets, showing consistent performance drops on non-standard dialectal inputs.

\emph{(ii) Multilingual and Non-English Benchmarks.} \citet{faisal2024dialectbench} present DialectBench, evaluating 281 language varieties across 40 language clusters and 10 NLP tasks. In Arabic, \citet{mousi-etal-2025-aradice} introduce AraDiCE, covering Modern Standard Arabic and three dialects across multiple tasks. \citet{altakrori2026dialectalarabicmmlu} adapt MMLU-style question answering into five Arabic dialects.

\noindent\textbf{Vietnamese Dialect studies in NLP.}
Vietnamese dialect processing has recently attracted increasing attention as regional variation challenges NLP systems trained mainly on standard, Northern-based Vietnamese. \citet{le2023centralnorthern} introduce a Central--Northern dialect transfer corpus and show that commercial translation systems struggle with Central dialect expressions, while monolingual Vietnamese models such as BARTpho outperform multilingual models for dialect transfer. Their fine-tuned transfer modules also improve downstream machine translation and text-image retrieval without retraining full systems. Extending this line of work, \citet{ta2026vidia2std} propose ViDia2Std, a nationwide dialect-to-standard corpus of over 13,000 sentence pairs from all 63 provinces, covering Central, Southern, and non-standard Northern varieties, and show that dialect normalization improves machine translation and sentiment analysis.

Prior work treats Vietnamese dialectal variation mainly as a preprocessing problem, showing that dialect-to-standard normalization can improve downstream systems. However, this does not reveal whether models are inherently robust to dialectal inputs: a model may perform well after normalization yet behave inconsistently on meaning-preserving dialectal variants. \emph{Our work instead focuses on dialectal robustness evaluation, asking whether LLMs produce stable and correct predictions across semantically equivalent Vietnamese dialect forms. This enables controlled analysis of performance drops across tasks, dialect regions, and model families.}
\section{Dataset Construction}
\label{sec:dataset}

\begin{figure*}[t]
    \centering
    \includegraphics[width=0.9\textwidth]{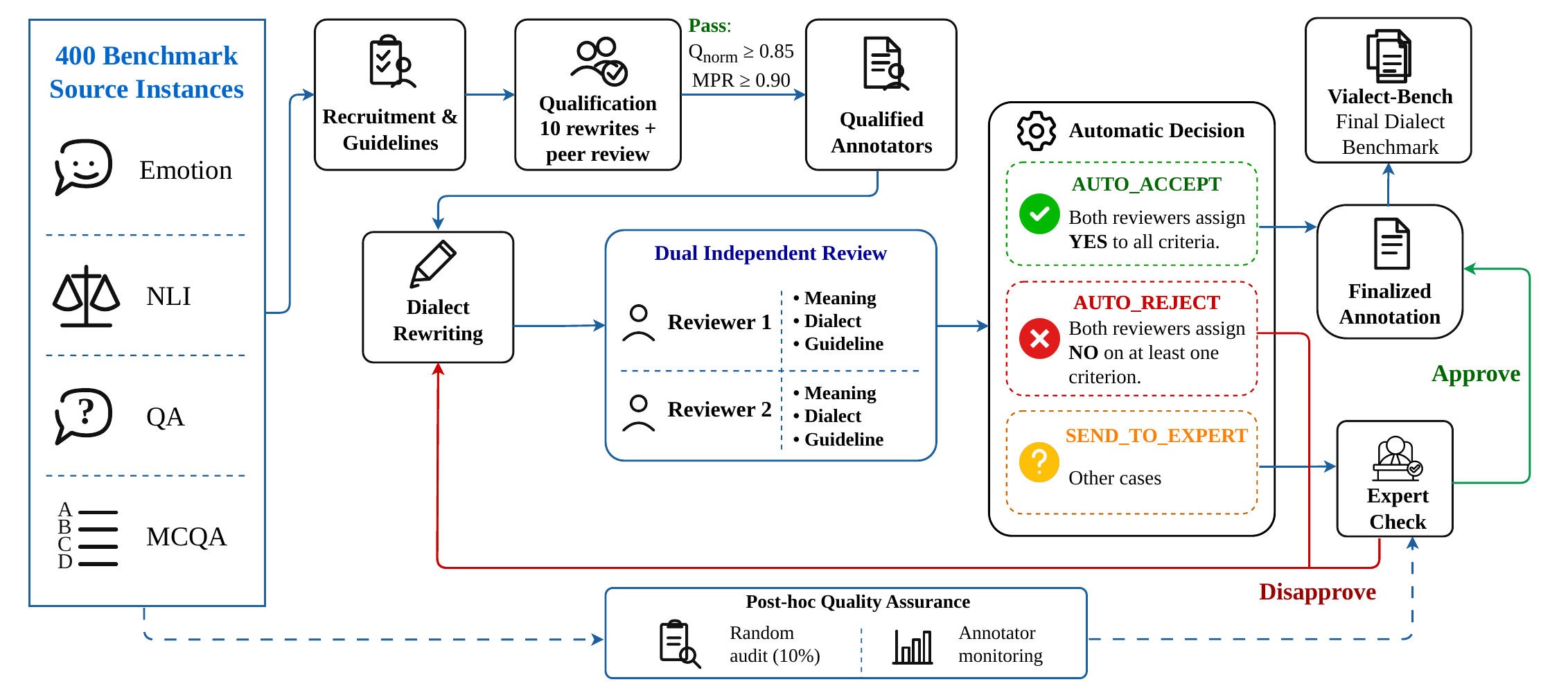}
    \caption{Overview of the annotation and quality-assurance pipeline for \textbf{VialectBench}.}
    \label{fig:dataset_pipeline}
\end{figure*}

\textbf{VialectBench} is a manually annotated benchmark comprising 400 standard Vietnamese source instances and 2,400 meaning-preserving dialectal rewrites across six Vietnamese dialect groups. Each source instance is paired with six regional variants. For each source instance, we keep the task label and task-defining evidence fixed while rewriting only the task-bearing input fields: the utterance for ER, the hypothesis for NLI, the question for QA, and both the question and answer options for MCQA. Premises, QA contexts, answer spans, and gold labels are preserved. For MCQA, the semantic identity of each option and the correct answer index are preserved, so that any prediction change reflects dialectal variation rather than a change in the underlying task. Together, these four tasks allow us to evaluate dialectal robustness beyond surface recognition, covering affective understanding, semantic inference, evidence grounding, and answer selection.


We first describe the data sources and sampling strategies in Section~\ref{subsec:data_source}, and then detail the annotation and quality assurance processes in Section~\ref{subsec:annotation}.
\subsection{Data Sourcing}
\label{subsec:data_source}
In this section, we introduce the selected datasets and their domains. We select these benchmarks to approximate real-world Vietnamese language-use scenarios across diverse domains.

\begin{table}[!b]
\centering
\footnotesize
\setlength{\tabcolsep}{4pt}
\renewcommand{\arraystretch}{1.08}
\begin{tabular*}{\columnwidth}{@{\extracolsep{\fill}}lrrr@{}}
\toprule
Criterion & Yes & Partial & No \\
\midrule
Semantic preservation & 72.1\% & 21.4\% & 6.5\% \\
Dialect authenticity  & 58.4\% & 36.6\% & 5.0\% \\
Guideline compliance  & 76.2\% & 22.3\% & 1.5\% \\
\bottomrule
\end{tabular*}
\caption{Dual-reviewer evaluation outcomes for all 2,400 dialectal rewrites, rated across three quality criteria.}
\label{tab:review_summary_combined}
\end{table}
\noindent\textbf{ER -- UIT-VSMEC.} \cite{ho2020emotionrecognitionvietnamesesocial} is a Vietnamese social-media emotion recognition dataset consisting of 6,927 human-annotated sentences labeled with six basic emotions and an additional other category.

\noindent\textbf{NLI -- ViANLI.} \cite{vanhuynh2025newbenchmarkdatasetmixtureofexperts} is a Vietnamese adversarial natural language inference dataset containing over 10,000 premise--hypothesis pairs constructed from more than 700 online news articles.

\noindent\textbf{QA -- ViQuAD.} \cite{Nguyen_2022} is a Vietnamese machine reading comprehension dataset consisting of over 23,000 human-generated question--answer pairs based on 5,109 passages from 174 Vietnamese Wikipedia articles.

\noindent\textbf{MCQA -- ViMMRC 2.0.} \cite{Luu_2025} is a Vietnamese multiple-choice reading comprehension dataset comprising 699 reading passages and 5,273 questions collected from literature materials for students from Grades 1 to 12.
 
Rather than relying on random sampling, we adopt a lexicon-guided selection strategy to increase the presence of dialect-sensitive phenomena. Candidate instances are identified using a curated Vietnamese dialect lexicon covering lexical, functional, pronominal, discourse-level, and idiomatic variation. We prioritize examples with richer dialectal trigger coverage while applying task-specific quality constraints, such as valid labels, answerable QA instances, bounded answer length, and well-formed MCQA options.

\subsection{Annotation and Quality Assurance}
\label{subsec:annotation}

\textbf{VialectBench} follows recent dialect robustness benchmarks that construct parallel inputs through human rewriting or human post-editing rather than relying solely on automatic perturbations~\citep{lin2025redial,mousi-etal-2025-aradice,gupta2025endive}. For Vietnamese, our design is also informed by prior dialect-to-standard resources~\citep{le2023centralnorthern,ta2026vidia2std}; however, unlike sentence-level normalization corpora, each dialectal rewrite in \textbf{VialectBench} is tied to a downstream task label and paired with a standard-control input for direct robustness evaluation.

The six groups cover Northern Vietnamese (PNB), North-Central I (PNT1; Thanh Hoa), North-Central II (PNT2; Nghe An--Ha Tinh), North-Central III (PNT3; Quang Binh--Quang Tri--Thua Thien Hue), South-Central (PNT4; Da Nang--Binh Thuan), and Southern Vietnamese (PNN; Ho Chi Minh City, Southeast Vietnam, and the Mekong Delta). This taxonomy is designed to cover major regional varieties while also distinguishing several Central Vietnamese subregions, where lexical, pronominal, and discourse-marker variation is especially salient. 


\begin{table}[!b]
\centering
\scriptsize
\setlength{\tabcolsep}{2.5pt}
\renewcommand{\arraystretch}{1.05}
\resizebox{\columnwidth}{!}{%
\begin{tabular}{lrrrr}
\toprule
Group & Standard PPL & Dialect PPL & PPL ratio & Length ratio \\
\midrule
\multicolumn{5}{l}{\textit{By dialect}} \\
PNB  & 93.178 & 82.909  & 0.926 & 0.995 \\
PNN  & 93.178 & 136.919 & 1.493 & 0.984 \\
PNT1 & 93.178 & 166.437 & 1.905 & 0.997 \\
PNT2 & 93.178 & 257.017 & 2.726 & 0.991 \\
PNT3 & 93.178 & 263.194 & $\textbf{3.334}$ & 0.989 \\
PNT4 & 93.178 & 158.199 & 1.903 & 0.986 \\
\midrule
\multicolumn{5}{l}{\textit{By task}} \\
MCQA    & 21.998  & 31.260  & 1.444 & 0.995 \\
NLI     & 132.429 & 239.129 & 2.244 & 0.998 \\
QA      & 47.171  & 103.105 & $\textbf{2.385}$ & 0.985 \\
Emotion & 171.115 & 336.290 & 2.118 & 0.982 \\
\midrule
All & 93.178 & 177.446 & $\textbf{2.048}$ & 0.990 \\
\bottomrule
\end{tabular}%
}
\caption{Dataset quality evaluated with \texttt{Qwen2.5-0.5B} as a fixed scorer. Standard and dialect PPL are averaged across sequences, while PPL and length ratios are averaged across matched pairs.}
\label{tab:intrinsic}
\end{table}
Figure~\ref{fig:dataset_pipeline} summarizes the annotation and quality-assurance workflow. Annotators are recruited from the six dialect regions and self-identified as native speakers or long-term residents of the corresponding region. This follows recent dialect benchmark construction practices that rely on speakers of the target variety to produce or validate dialectal variants~\cite{lin2025redial}. Before entering the main workflow, annotators complete a qualification stage consisting of dialect rewriting and peer review. Annotators pass qualification when two criteria are met: their combined dialect-naturalness and guideline-compliance score $Q_{\mathrm{norm}}$ (the average of reviewer ratings on a 0--1 scale across those two dimensions) is at least 0.85, and their meaning-preservation rate over the qualification rewrites is at least 0.90.

Each submitted rewrite is checked by two independent reviewers along three dimensions: semantic preservation, dialect authenticity, and guideline compliance. 

We use dual review and adjudication to reduce individual annotator bias and to support reliable corpus construction, following standard practice in linguistic annotation and inter-coder agreement analysis~\cite{artstein-poesio-2008-survey}.


Table~\ref{tab:review_summary_combined} reports the aggregated outcomes from the dual-reviewer evaluation across all 2,400 rewrites. Guideline compliance is the strongest dimension, with 76.2\% of rewrites fully passing and only 1.5\% failing. Semantic preservation is similarly high at 72.1\%, though 6.5\% of rewrites were flagged for meaning loss. Dialect authenticity shows the most partial ratings (36.6\%), reflecting the difficulty of judging fine-grained regional naturalness.

\begin{table*}[!t]
\centering
\scriptsize
\setlength{\tabcolsep}{2.7pt}
\renewcommand{\arraystretch}{1.05}
\resizebox{\textwidth}{!}{%
\begin{tabular}{llrrrrrrrrrr}
\toprule
Task & Variant & Qwen2.5 0.5B & Qwen2.5 3B & Qwen2.5 7B & Llama 3.1 8B & Mistral 7B & Gemma 2 9B & Gemma 3 4B & Vistral 7B & SeaLLM 7B & GPT-4o \\
\midrule
MCQA & Standard & {\footnotesize 68.0} & {\footnotesize 92.0} & {\footnotesize 91.0} & {\footnotesize 87.0} & {\footnotesize 73.0} & {\footnotesize 94.0} & {\footnotesize 89.0} & {\footnotesize 88.0} & {\footnotesize 91.0} & {\footnotesize 94.0} \\
 & PNB & {\footnotesize 63.0} & {\footnotesize 92.0} & {\footnotesize 90.0} & {\footnotesize 87.0} & {\footnotesize 75.0} & {\footnotesize 94.0} & {\footnotesize 90.0} & {\footnotesize 88.0} & {\footnotesize 91.0} & {\footnotesize 94.0} \\
 & PNN & {\footnotesize 63.0} & {\footnotesize 90.0} & {\footnotesize 89.0} & {\footnotesize 85.0} & {\footnotesize 69.0} & {\footnotesize 95.0} & {\footnotesize 88.0} & {\footnotesize 88.0} & {\footnotesize 88.0} & {\footnotesize 94.0} \\
 & PNT1 & {\footnotesize 57.0} & {\footnotesize 87.0} & {\footnotesize 88.0} & {\footnotesize 81.0} & {\footnotesize 72.0} & {\footnotesize 93.0} & {\footnotesize 88.0} & {\footnotesize 87.0} & {\footnotesize 85.0} & {\footnotesize 94.0} \\
 & PNT2 & {\footnotesize 62.0} & {\footnotesize 87.0} & {\footnotesize 86.0} & {\footnotesize 80.0} & {\footnotesize 70.0} & {\footnotesize 96.0} & {\footnotesize 86.0} & {\footnotesize 85.0} & {\footnotesize 88.0} & {\footnotesize 94.0} \\
 & PNT3 & {\footnotesize 57.0} & {\footnotesize 84.0} & {\footnotesize 86.0} & {\footnotesize 81.0} & {\footnotesize 66.0} & {\footnotesize 92.0} & {\footnotesize 88.0} & {\footnotesize 87.0} & {\footnotesize 90.0} & {\footnotesize 93.0} \\
 & PNT4 & {\footnotesize 61.0} & {\footnotesize 88.0} & {\footnotesize 88.0} & {\footnotesize 85.0} & {\footnotesize 72.0} & {\footnotesize 95.0} & {\footnotesize 87.0} & {\footnotesize 88.0} & {\footnotesize 88.0} & {\footnotesize 94.0} \\
\midrule
NLI & Standard & {\footnotesize 33.0} & {\footnotesize 55.0} & {\footnotesize 64.0} & {\footnotesize 48.0} & {\footnotesize 47.0} & {\footnotesize 63.0} & {\footnotesize 55.0} & {\footnotesize 19.0} & {\footnotesize 46.0} & {\footnotesize 74.0} \\
 & PNB & {\footnotesize 33.0} & {\footnotesize 56.0} & {\footnotesize 66.0} & {\footnotesize 48.0} & {\footnotesize 49.0} & {\footnotesize 65.0} & {\footnotesize 57.0} & {\footnotesize 21.0} & {\footnotesize 48.0} & {\footnotesize 76.0} \\
 & PNN & {\footnotesize 33.0} & {\footnotesize 55.0} & {\footnotesize 59.0} & {\footnotesize 41.0} & {\footnotesize 45.0} & {\footnotesize 60.0} & {\footnotesize 56.0} & {\footnotesize 17.0} & {\footnotesize 45.0} & {\footnotesize 69.0} \\
 & PNT1 & {\footnotesize 33.0} & {\footnotesize 54.0} & {\footnotesize 60.0} & {\footnotesize 44.0} & {\footnotesize 51.0} & {\footnotesize 62.0} & {\footnotesize 50.0} & {\footnotesize 19.0} & {\footnotesize 44.0} & {\footnotesize 71.0} \\
 & PNT2 & {\footnotesize 33.0} & {\footnotesize 56.0} & {\footnotesize 61.0} & {\footnotesize 47.0} & {\footnotesize 48.0} & {\footnotesize 60.0} & {\footnotesize 54.0} & {\footnotesize 18.0} & {\footnotesize 43.0} & {\footnotesize 70.0} \\
 & PNT3 & {\footnotesize 33.0} & {\footnotesize 50.0} & {\footnotesize 59.0} & {\footnotesize 45.0} & {\footnotesize 44.0} & {\footnotesize 57.0} & {\footnotesize 45.0} & {\footnotesize 18.0} & {\footnotesize 42.0} & {\footnotesize 69.0} \\
 & PNT4 & {\footnotesize 33.0} & {\footnotesize 55.0} & {\footnotesize 62.0} & {\footnotesize 48.0} & {\footnotesize 50.0} & {\footnotesize 62.0} & {\footnotesize 51.0} & {\footnotesize 18.0} & {\footnotesize 47.0} & {\footnotesize 72.0} \\
\midrule
QA & Standard & {\footnotesize 38.9} & {\footnotesize 69.2} & {\footnotesize 74.7} & {\footnotesize 79.4} & {\footnotesize 54.3} & {\footnotesize 84.1} & {\footnotesize 79.6} & {\footnotesize 66.2} & {\footnotesize 74.3} & {\footnotesize 83.2} \\
 & PNB & {\footnotesize 39.5} & {\footnotesize 68.0} & {\footnotesize 73.2} & {\footnotesize 78.6} & {\footnotesize 56.0} & {\footnotesize 82.8} & {\footnotesize 79.9} & {\footnotesize 65.0} & {\footnotesize 74.2} & {\footnotesize 84.6} \\
 & PNN & {\footnotesize 37.6} & {\footnotesize 65.5} & {\footnotesize 71.4} & {\footnotesize 78.5} & {\footnotesize 46.2} & {\footnotesize 83.0} & {\footnotesize 79.8} & {\footnotesize 62.1} & {\footnotesize 72.4} & {\footnotesize 84.1} \\
 & PNT1 & {\footnotesize 33.4} & {\footnotesize 65.8} & {\footnotesize 70.2} & {\footnotesize 75.4} & {\footnotesize 48.1} & {\footnotesize 82.8} & {\footnotesize 74.1} & {\footnotesize 63.1} & {\footnotesize 67.0} & {\footnotesize 83.7} \\
 & PNT2 & {\footnotesize 18.7} & {\footnotesize 59.1} & {\footnotesize 68.7} & {\footnotesize 75.1} & {\footnotesize 30.5} & {\footnotesize 82.0} & {\footnotesize 73.8} & {\footnotesize 60.3} & {\footnotesize 66.2} & {\footnotesize 86.2} \\
 & PNT3 & {\footnotesize 17.1} & {\footnotesize 57.6} & {\footnotesize 67.3} & {\footnotesize 74.5} & {\footnotesize 34.4} & {\footnotesize 80.1} & {\footnotesize 74.5} & {\footnotesize 57.7} & {\footnotesize 61.5} & {\footnotesize 81.6} \\
 & PNT4 & {\footnotesize 32.6} & {\footnotesize 60.3} & {\footnotesize 68.0} & {\footnotesize 76.9} & {\footnotesize 37.1} & {\footnotesize 80.0} & {\footnotesize 71.8} & {\footnotesize 60.6} & {\footnotesize 68.1} & {\footnotesize 82.2} \\
\midrule
Emotion & Standard & {\footnotesize 24.0} & {\footnotesize 51.0} & {\footnotesize 38.0} & {\footnotesize 43.0} & {\footnotesize 37.0} & {\footnotesize 49.0} & {\footnotesize 43.0} & {\footnotesize 32.0} & {\footnotesize 37.0} & {\footnotesize 60.0} \\
 & PNB & {\footnotesize 23.0} & {\footnotesize 49.0} & {\footnotesize 41.0} & {\footnotesize 47.0} & {\footnotesize 40.0} & {\footnotesize 50.0} & {\footnotesize 45.0} & {\footnotesize 32.0} & {\footnotesize 38.0} & {\footnotesize 56.0} \\
 & PNN & {\footnotesize 24.0} & {\footnotesize 51.0} & {\footnotesize 38.0} & {\footnotesize 44.0} & {\footnotesize 36.0} & {\footnotesize 45.0} & {\footnotesize 41.0} & {\footnotesize 33.0} & {\footnotesize 39.0} & {\footnotesize 56.0} \\
 & PNT1 & {\footnotesize 24.0} & {\footnotesize 50.0} & {\footnotesize 38.0} & {\footnotesize 46.0} & {\footnotesize 37.0} & {\footnotesize 45.0} & {\footnotesize 42.0} & {\footnotesize 34.0} & {\footnotesize 39.0} & {\footnotesize 61.0} \\
 & PNT2 & {\footnotesize 22.0} & {\footnotesize 38.0} & {\footnotesize 35.0} & {\footnotesize 30.0} & {\footnotesize 33.0} & {\footnotesize 36.0} & {\footnotesize 39.0} & {\footnotesize 31.0} & {\footnotesize 35.0} & {\footnotesize 56.0} \\
 & PNT3 & {\footnotesize 20.0} & {\footnotesize 34.0} & {\footnotesize 33.0} & {\footnotesize 30.0} & {\footnotesize 29.0} & {\footnotesize 40.0} & {\footnotesize 41.0} & {\footnotesize 29.0} & {\footnotesize 34.0} & {\footnotesize 60.0} \\
 & PNT4 & {\footnotesize 22.0} & {\footnotesize 46.0} & {\footnotesize 44.0} & {\footnotesize 42.0} & {\footnotesize 39.0} & {\footnotesize 43.0} & {\footnotesize 42.0} & {\footnotesize 33.0} & {\footnotesize 37.0} & {\footnotesize 62.0} \\
\midrule
Mean & Standard & {\footnotesize 41.0} & {\footnotesize 66.8} & {\footnotesize 66.9} & {\footnotesize 64.3} & {\footnotesize 52.8} & {\footnotesize 72.5} & {\footnotesize 66.6} & {\footnotesize 51.3} & {\footnotesize 62.1} & {\footnotesize 77.8} \\
 & PNB & {\footnotesize 39.6}{\tiny\,$\downarrow$-1.4} & {\footnotesize 66.3}{\tiny\,$\downarrow$-0.5} & {\footnotesize 67.5}{\tiny\,$\uparrow$+0.6} & {\footnotesize 65.1}{\tiny\,$\uparrow$+0.8} & {\footnotesize 55.0}{\tiny\,$\uparrow$+2.2} & {\footnotesize 73.0}{\tiny\,$\uparrow$+0.4} & {\footnotesize 68.0}{\tiny\,$\uparrow$+1.3} & {\footnotesize 51.5}{\tiny\,$\uparrow$+0.2} & {\footnotesize 62.8}{\tiny\,$\uparrow$+0.7} & {\footnotesize 77.7}{\tiny\,$\downarrow$-0.2} \\
 & PNN & {\footnotesize 39.4}{\tiny\,$\downarrow$-1.6} & {\footnotesize 65.4}{\tiny\,$\downarrow$-1.4} & {\footnotesize 64.4}{\tiny\,$\downarrow$-2.6} & {\footnotesize 62.1}{\tiny\,$\downarrow$-2.2} & {\footnotesize 49.0}{\tiny\,$\downarrow$-3.8} & {\footnotesize 70.7}{\tiny\,$\downarrow$-1.8} & {\footnotesize 66.2}{\tiny\,$\downarrow$-0.4} & {\footnotesize 50.0}{\tiny\,$\downarrow$-1.3} & {\footnotesize 61.1}{\tiny\,$\downarrow$-1.0} & {\footnotesize 75.8}{\tiny\,$\downarrow$-2.0} \\
 & PNT1 & {\footnotesize 36.8}{\tiny\,$\downarrow$-4.1} & {\footnotesize 64.2}{\tiny\,$\downarrow$-2.6} & {\footnotesize 64.0}{\tiny\,$\downarrow$-2.9} & {\footnotesize 61.6}{\tiny\,$\downarrow$-2.7} & {\footnotesize 52.0}{\tiny\,$\downarrow$-0.8} & {\footnotesize 70.7}{\tiny\,$\downarrow$-1.8} & {\footnotesize 63.5}{\tiny\,$\downarrow$-3.1} & {\footnotesize 50.8}{\tiny\,$\downarrow$-0.5} & {\footnotesize 58.7}{\tiny\,$\downarrow$-3.3} & {\footnotesize 77.4}{\tiny\,$\downarrow$-0.4} \\
 & PNT2 & {\footnotesize 33.9}{\tiny\,$\downarrow$-7.1} & {\footnotesize 60.0}{\tiny\,$\downarrow$-6.8} & {\footnotesize 62.7}{\tiny\,$\downarrow$-4.2} & {\footnotesize 58.0}{\tiny\,$\downarrow$-6.3} & {\footnotesize 45.4}{\tiny\,$\downarrow$-7.4} & {\footnotesize 68.5}{\tiny\,$\downarrow$-4.0} & {\footnotesize 63.2}{\tiny\,$\downarrow$-3.4} & {\footnotesize 48.6}{\tiny\,$\downarrow$-2.7} & {\footnotesize 58.0}{\tiny\,$\downarrow$-4.0} & {\footnotesize 76.6}{\tiny\,$\downarrow$-1.3} \\
 & PNT3 & {\footnotesize 31.8}{\tiny\,$\downarrow$-9.2} & {\footnotesize 56.4}{\tiny\,$\downarrow$-10.4} & {\footnotesize 61.3}{\tiny\,$\downarrow$-5.6} & {\footnotesize 57.6}{\tiny\,$\downarrow$-6.7} & {\footnotesize 43.4}{\tiny\,$\downarrow$-9.5} & {\footnotesize 67.3}{\tiny\,$\downarrow$-5.3} & {\footnotesize 62.1}{\tiny\,$\downarrow$-4.5} & {\footnotesize 47.9}{\tiny\,$\downarrow$-3.4} & {\footnotesize 56.9}{\tiny\,$\downarrow$-5.2} & {\footnotesize 75.9}{\tiny\,$\downarrow$-1.9} \\
 & PNT4 & {\footnotesize 37.1}{\tiny\,$\downarrow$-3.8} & {\footnotesize 62.3}{\tiny\,$\downarrow$-4.5} & {\footnotesize 65.5}{\tiny\,$\downarrow$-1.4} & {\footnotesize 63.0}{\tiny\,$\downarrow$-1.4} & {\footnotesize 49.5}{\tiny\,$\downarrow$-3.3} & {\footnotesize 70.0}{\tiny\,$\downarrow$-2.5} & {\footnotesize 62.9}{\tiny\,$\downarrow$-3.7} & {\footnotesize 49.9}{\tiny\,$\downarrow$-1.4} & {\footnotesize 60.0}{\tiny\,$\downarrow$-2.1} & {\footnotesize 77.5}{\tiny\,$\downarrow$-0.3} \\
\bottomrule
\end{tabular}%
}
\caption{\textbf{Task-by-dialect direct-prompting performance (\%)}. Mean rows report the average across the four tasks, with the adjacent values showing the gap from each model's Standard Vietnamese score.}
\label{tab:task-dialect-performance-matrix}
\end{table*}

\subsection{Dataset Evaluation}
\label{sec:intrinsic}

This evaluation characterizes the finalized \textbf{VialectBench} before any downstream task evaluation. We use pretrained \texttt{Qwen2.5-0.5B}~\citep{qwen2025qwen25technicalreport} as a frozen causal-language-model scorer to measure how much each dialectal rewrite shifts the model's likelihood relative to its matched Standard Vietnamese control. We score only the task-bearing fields that are rewritten: the utterance for ER, the hypothesis for NLI, the question for QA, and the question together with its answer options for MCQA. For NLI, the Standard comparison is the original hypothesis rather than the unchanged premise. We report the within-pair perplexity ratio as the primary relative measure, along with the whitespace-token length ratio to verify that rewrites remain closely matched in length.

\noindent\textbf{Findings.}
Table~\ref{tab:intrinsic} shows a clear likelihood shift in the finalized dataset. Across all 2,400 matched pairs, the mean within-pair PPL ratio is $2.048$, while the mean length ratio is $0.990$, indicating that the rewrites remain closely matched to their Standard controls in length while exhibiting higher perplexity under the reference scorer.

The shift is strongly dialect-dependent. PNT3 and PNT2 have the largest PPL ratios, at $3.334$ and $2.726$, respectively, followed by PNT1 ($1.905$), PNT4 ($1.903$), and PNN ($1.493$). PNB is the only group below 1 ($0.926$), indicating that its rewrites are slightly more probable than their matched Standard forms under this scorer. Across tasks, QA exhibits the largest ratio ($2.385$), followed by NLI ($2.244$), ER ($2.118$), and MCQA ($1.444$).

These results establish, before downstream evaluation, that \textbf{VialectBench} contains a measurable and heterogeneous model-relative surface-form shift. They do not rank dialects by linguistic quality or imply that a larger likelihood shift must cause a larger task-performance drop; Section~\ref{sec:experiments} separately evaluates whether model decisions remain stable under these matched rewrites.

\section{Experiments}
\label{sec:experiments} 
\subsection{Experimental Setting}
We evaluate LLM robustness on \textbf{VialectBench}, which provides six meaning-preserving dialectal variants for each standard Vietnamese instance across four task families.

\noindent\textbf{Models.} We evaluate ten instruction-tuned models: nine open-source models and one proprietary model.\ding{182} The open-source models include three Qwen2.5-Instruct
variants (0.5B, 3B, and 7B)~\citep{qwen2025qwen25technicalreport},
Llama-3.1-8B-Instruct~\citep{grattafiori2024llama3herdmodels},
Mistral-7B-Instruct-v0.3~\citep{jiang2023mistral7b},
Gemma-2-9B-IT~\citep{gemmateam2024gemma2improvingopen}, and
Gemma-3-4B-IT~\citep{gemmateam2025gemma3technicalreport} as general-purpose models, together with Vistral-7B-Chat~\cite{chien2023vistral} and
SeaLLMs-v3-7B-Chat~\cite{zhang-etal-2025-seallms} as Vietnamese- and Southeast
Asian-adapted models, respectively.\footnote{
All open-source models are accessed through Hugging Face.
}
\ding{183} We additionally evaluate the closed-source model
GPT-4o.\footnote{
OpenAI API, \url{https://platform.openai.com/docs/api-reference};
we use the fixed \texttt{gpt-4o-2024-08-06} API snapshot.
Accessed on 5th July 2026.
}. This selection enables comparisons across model scales and families, while also examining whether regional language adaptation or stronger general-purpose capabilities improve robustness to Vietnamese dialectal variation.

\begin{figure*}[t]
    \centering
    \includegraphics[width=\textwidth]
    {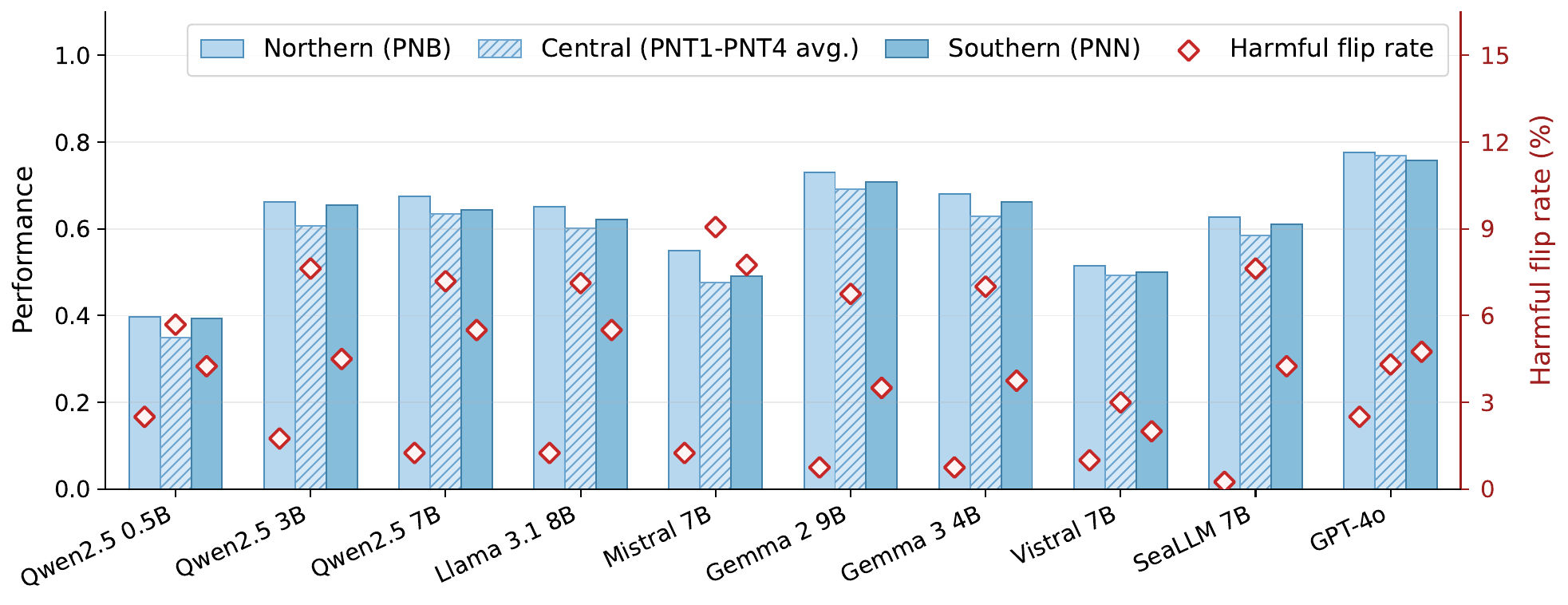}
    \caption{Regional performance (left axis) and harmful-flip rate (right axis) by model.}
    \label{fig:model_performance_flip}
\end{figure*}

\noindent\textbf{Implementation.} We use direct prompting as the primary inference setting for all models. Each benchmark instance is evaluated in its Standard Vietnamese form and across all corresponding dialectal variants using the same task-specific instruction. Models are required to return a JSON-formatted response, which is subsequently parsed into a task-specific prediction. We use deterministic decoding throughout: for Hugging Face models, we disable sampling by setting
\texttt{do\_sample=False}, whereas GPT-4o is queried with
\texttt{temperature=0}. These deterministic decoding settings ensure that
differences between Standard Vietnamese and dialectal variants primarily
reflect input variation rather than stochasticity during generation.

\noindent\textbf{Evaluation.}
We evaluate MCQA, NLI, and ER using accuracy, and extractive QA using token-level F1. We additionally report the harmful-flip rate (HFR), defined as the percentage of matched examples answered correctly in Standard Vietnamese but incorrectly in dialectal forms. For QA, correctness in the flip analysis is determined using exact match, while token-level F1 remains the primary performance metric. 
For MCQA, NLI, and ER, we further measure confidence erosion among examples predicted correctly in both their Standard and dialectal forms. GPT-4o is excluded from this analysis because comparable label probabilities are unavailable.

\subsection{Experimental Results}
We report task performance on \emph{VialectBench} in Table~\ref{tab:task-dialect-performance-matrix} and Table~\ref{tab:task-dialect-delta}, and summarize the main results of our experiments through these findings.

\noindent\textbf{Dialect brittleness is present across all evaluated models.}
We define the dialect performance gap as the absolute
difference between a model's Standard Vietnamese score and
its score on a dialectal variant. As shown in
Table~\ref{tab:task-dialect-performance-matrix}, PNT3 produces
the largest degradation for nine of the ten evaluated models.
The most severe cases are Qwen2.5-3B, Mistral-7B, and
Qwen2.5-0.5B, whose performance under PNT3 decreases by
$10.4\%$, $9.5\%$, and $9.2\%$, respectively. Because these
large drops occur across different model families, dialect
brittleness is not confined to a particular architecture or
model series. At the same time, dialectal variation is not
uniformly harmful: PNB occasionally improves performance
relative to Standard Vietnamese for some models. This
suggests that certain regional reformulations may better match
a model's learned lexical or syntactic preferences.

Importantly, absolute capability and dialect stability are
distinct. GPT-4o is both the strongest and the most stable
model: it achieves the highest Standard Vietnamese score of
$77.8\%$ and has an average dialect gap of only approximately
$1.0\%$. In contrast, Vistral-7B also exhibits a relatively
small average gap of $1.5\%$, but its absolute performance
remains substantially lower, ranging from $47.9\%$ to
$51.5\%$ across dialect groups. Thus, a small dialect gap
does not necessarily indicate strong robustness; it may also
reflect consistently low performance across both Standard
and dialectal inputs. Likewise, occasional improvements under
PNB should not be interpreted as dialect invariance, since
the same models can still suffer substantial losses under
other regional variants.

\begin{table}[!t]
\centering
\footnotesize
\renewcommand{\arraystretch}{1.05}
\begin{tabular*}{\columnwidth}
{@{\extracolsep{\fill}}lrrrrrr@{}}
\toprule
Task & PNB & PNN & PNT1 & PNT2 & PNT3 & PNT4 \\
\midrule
MCQA
& 0.30 & 1.80 & 3.50 & 3.30 & 4.30 & 2.10 \\
NLI
& -1.50 & 2.40 & 1.60 & 1.40 & 4.20 & 0.60 \\
QA
& 0.21 & 2.33 & 4.02 & 8.31
& \textbf{9.76} & 6.65 \\
Emotion
& -0.70 & 0.70 & -0.20 & 5.90 & 6.40 & 0.40 \\
\midrule
Mean
& -0.42 & 1.81 & 2.23 & 4.73
& \textbf{6.17} & 2.44 \\
\bottomrule
\end{tabular*}
\caption{Mean performance gap
$S_{\mathrm{Std}}-S_d$ (\%) by task and dialect, averaged
across ten models. Positive values indicate degradation.}
\label{tab:task-dialect-delta}
\end{table}

\noindent\textbf{Dialectal variation affects all task families, but QA shows the broadest vulnerability across Central Vietnamese.}
Tables~\ref{tab:task-dialect-performance-matrix} and
\ref{tab:task-dialect-delta} show that all four tasks degrade
on average, with QA exhibiting the largest mean gap
($5.21\%$), followed by MCQA ($2.55\%$), ER ($2.08\%$),
and NLI ($1.45\%$). More importantly, QA is the only task
with substantial degradation across all four Central variants:
$4.02\%$, $8.31\%$, $9.76\%$, and $6.65\%$ for
PNT1--PNT4, respectively. In contrast, ER and NLI are
affected more selectively, with their largest losses
concentrated in PNT2 and PNT3. Because QA rewrites the
question while preserving the Standard Vietnamese context,
this pattern suggests that dialectal variation particularly
disrupts the alignment between a regional query and its
supporting evidence.

NLI has the smallest mean degradation, yet it still drops by
$4.20\%$ under PNT3, while improving by $1.50\%$ under
PNB. Likewise, ER changes little under PNB, PNN, PNT1,
and PNT4, but declines sharply under PNT2 and PNT3 by
$5.90\%$ and $6.40\%$. MCQA follows the same regional
pattern, with its largest loss also occurring under PNT3
($4.30\%$). Figure~\ref{fig:model_performance_flip}
reinforces this asymmetry: the aggregated Central group
generally yields both lower performance and higher
harmful-flip rates than PNB and PNN. These results indicate
that dialect robustness depends jointly on the task operation
and the regional variant, and that averaging across dialects
can obscure the concentrated brittleness caused by PNT2 and
PNT3.

\noindent\textbf{Central dialects produce both lower
performance and more consequential prediction failures.}
Figure~\ref{fig:model_performance_flip} shows a consistent
regional asymmetry across model families: the aggregated
Central group generally yields the lowest performance and
the highest harmful-flip rate, whereas PNB is typically the
least disruptive. This indicates that the Central effect is not
limited to small changes in aggregate scores; it more often
converts predictions that are correct in Standard Vietnamese
into incorrect dialectal predictions. The figure also shows
that absolute capability and stability must be considered
jointly. GPT-4o maintains both high regional performance
and relatively low harmful-flip rates, while Vistral-7B appears
stable across regions only at a much lower performance level.
Conversely, Mistral-7B exhibits both a pronounced Central
performance decline and a high harmful-flip rate. Thus,
similar average regional gaps can correspond to substantially
different levels of practical reliability.

\section{Limitations}
Although \textbf{VialectBench} evaluates controlled written dialect variation, several limitations remain. First, the lexicon-guided sampling strategy intentionally enriches dialect-sensitive phenomena and may therefore over-represent particular lexical or functional patterns relative to naturally occurring Vietnamese. Second, the intrinsic evaluation also relies on a single reference model and tokenizer, so the observed perplexity ordering may be scorer-specific. Finally, we do not evaluate lightweight mitigation strategies, such as dialect-to-Standard normalization or normalization-aware prompting, and therefore cannot determine how much of the observed robustness gap is recoverable without retraining.

\section{Conclusion}
This paper presented \textbf{VialectBench}, a controlled benchmark for evaluating the robustness of LLMs when prompted with six meaning-preserving rewritten Vietnamese regional varieties. The benchmark contains 400 source instances and 2,400 human-written dialectal variants across ER, NLI, QA, and MCQA. Evaluation of ten instruction-tuned models showed that dialectal variation consistently affects performance, and the effect is strongly region-dependent: PNT3 and PNT2 produce the largest average losses, while PNB yields a slight improvement. QA is the most vulnerable task, and the pooled Central group produces the highest harmful-flip rate. Overall, the results show that strong performance on Standard Vietnamese alone is insufficient to establish robustness across regional Vietnamese.

\section{Acknowledgement}
This research was supported by SEAS -- Summer in Engineering and Applied Sciences. We sincerely thank the program for providing the opportunity, resources, and collaborative environment that made this project possible. We are also grateful to Dang Thi Ngoc Anh, Nguyen Ha Phuong, Nguyen Nhat Anh, Nguyen Quang Ly, and Nguyen Thanh Tu for their valuable discussions, feedback, and support throughout the summer school.
\bibliography{ref}

@misc{zhao2026surveylargelanguagemodels,
      title={A Survey of Large Language Models}, 
      author={Wayne Xin Zhao and Kun Zhou and Junyi Li and Tianyi Tang and Xiaolei Wang and Yupeng Hou and Yingqian Min and Beichen Zhang and Junjie Zhang and Zican Dong and Yifan Du and Chen Yang and Yushuo Chen and Zhipeng Chen and Jinhao Jiang and Ruiyang Ren and Yifan Li and Xinyu Tang and Zikang Liu and Peiyu Liu and Jian-Yun Nie and Ji-Rong Wen},
      year={2026},
      eprint={2303.18223},
      archivePrefix={arXiv},
      primaryClass={cs.CL},
      url={https://arxiv.org/abs/2303.18223}, 
}

@inproceedings{zhu2024promptrobustevaluatingrobustnesslarge,
author = {Zhu, Kaijie and Wang, Jindong and Zhou, Jiaheng and Wang, Zichen and Chen, Hao and Wang, Yidong and Yang, Linyi and Ye, Wei and Zhang, Yue and Gong, Neil and Xie, Xing},
title = {PromptRobust: Towards Evaluating the Robustness of Large Language Models on Adversarial Prompts},
year = {2024},
isbn = {9798400712098},
publisher = {Association for Computing Machinery},
address = {New York, NY, USA},
url = {https://doi.org/10.1145/3689217.3690621},
doi = {10.1145/3689217.3690621},
booktitle = {Proceedings of the 1st ACM Workshop on Large AI Systems and Models with Privacy and Safety Analysis},
pages = {57–68},
numpages = {12},
location = {Salt Lake City, UT, USA},
series = {LAMPS '24}
}

@article{hofmann2024aigenerates,
    title = {AI generates covertly racist decisions about people based on their dialect},
    author = {Hofmann, Valentin and Kalluri, Pratyusha Ria and Jurafsky, Dan and King, Sharese},
    journal = {Nature},
    year = {2024},
    volume = {633},
    number = {8028},
    pages = {147--154},
    doi = {10.1038/s41586-024-07856-5}
}

@article{Nguyen_2022,
   title={VLSP 2021-ViMRC Challenge: Vietnamese Machine Reading Comprehension},
   volume={38},
   ISSN={2615-9260},
   url={http://dx.doi.org/10.25073/2588-1086/vnucsce.340},
   DOI={10.25073/2588-1086/vnucsce.340},
   number={2},
   journal={VNU Journal of Science: Computer Science and Communication Engineering},
   publisher={Vietnam National University Journal of Science},
   author={Nguyen, Kiet and Tran, Son Quoc and Nguyen, Luan Thanh and Huynh, Tin Van and Luu, Son Thanh and Nguyen, Ngan Luu-Thuy},
   year={2022},
   month=Dec }

@article{Luu_2025,
   title={ViMMRC 2.0 — Enhancing Machine Reading Comprehension on Vietnamese Literature Text},
   volume={35},
   ISSN={2424-791X},
   url={http://dx.doi.org/10.1142/S2717554525500043},
   DOI={10.1142/s2717554525500043},
   number={03},
   journal={International Journal of Asian Language Processing},
   publisher={World Scientific Pub Co Pte Ltd},
   author={Luu, Son T. and Hoang, Khoi Trong and Pham, Tuong Quang and Nguyen, Kiet Van and Nguyen, Ngan Luu-Thuy},
   year={2025},
   month=July }

@article{vanhuynh2025newbenchmarkdatasetmixtureofexperts,
title = {A new benchmark dataset and mixture-of-experts language models for adversarial natural language inference in Vietnamese},
journal = {Expert Systems with Applications},
volume = {306},
pages = {130109},
year = {2026},
issn = {0957-4174},
doi = {https://doi.org/10.1016/j.eswa.2025.130109},
url = {https://www.sciencedirect.com/science/article/pii/S095741742503725X},
author = {Tin {Van Huynh} and Kiet {Van Nguyen} and Ngan {Luu-Thuy Nguyen}}
}

@InProceedings{ho2020emotionrecognitionvietnamesesocial,
author="Ho, Vong Anh
and Nguyen, Duong Huynh-Cong
and Nguyen, Danh Hoang
and Pham, Linh Thi-Van
and Nguyen, Duc-Vu
and Nguyen, Kiet Van
and Nguyen, Ngan Luu-Thuy",
editor="Nguyen, Le-Minh
and Phan, Xuan-Hieu
and Hasida, K{\^o}iti
and Tojo, Satoshi",
title="Emotion Recognition for Vietnamese Social Media Text",
booktitle="Computational Linguistics",
year="2020",
publisher="Springer Singapore",
address="Singapore",
pages="319--333",
isbn="978-981-15-6168-9"
}

@inproceedings{bui-etal-2025-vmlu,
    title = "{VMLU} Benchmarks: A comprehensive benchmark toolkit for {V}ietnamese {LLM}s",
    author = "Bui, Cuc Thi  and
      Son, Nguyen Truong  and
      Trang, Truong Van  and
      Phung, Lam Viet  and
      Huy, Pham Nhut  and
      Le, Hoang Anh  and
      Van, Quoc Huu  and
      Do, Phong Nguyen-Thuan  and
      Truc, Van Le Tran  and
      Chau, Duc Thanh  and
      Nguyen, Le-Minh",
    editor = "Che, Wanxiang  and
      Nabende, Joyce  and
      Shutova, Ekaterina  and
      Pilehvar, Mohammad Taher",
    booktitle = "Proceedings of the 63rd Annual Meeting of the Association for Computational Linguistics (Volume 1: Long Papers)",
    month = jul,
    year = "2025",
    address = "Vienna, Austria",
    publisher = "Association for Computational Linguistics",
    url = "https://aclanthology.org/2025.acl-long.563/",
    doi = "10.18653/v1/2025.acl-long.563",
    pages = "11495--11515",
    ISBN = "979-8-89176-251-0"
}

@inproceedings{le2023centralnorthern,
    title = "A Parallel Corpus for {V}ietnamese Central-Northern Dialect Text Transfer",
    author = "Le, Thang  and
      Luu, Anh",
    editor = "Bouamor, Houda  and
      Pino, Juan  and
      Bali, Kalika",
    booktitle = "Findings of the Association for Computational Linguistics: EMNLP 2023",
    month = dec,
    year = "2023",
    address = "Singapore",
    publisher = "Association for Computational Linguistics",
    url = "https://aclanthology.org/2023.findings-emnlp.925/",
    doi = "10.18653/v1/2023.findings-emnlp.925",
    pages = "13839--13855"
}

@article{ta2026vidia2std,
  title={ViDia2Std: A Parallel Corpus and Methods for Low-Resource Vietnamese Dialect-to-Standard Translation},
  volume={40},
  url={https://ojs.aaai.org/index.php/AAAI/article/view/40247},
  DOI={10.1609/aaai.v40i36.40247},
  abstractNote={Vietnamese exhibits extensive dialectal variation, posing challenges for NLP systems trained predominantly on standard Vietnamese. Such systems often underperform on dialectal inputs, especially from underrepresented Central and Southern regions. Previous work on dialect normalization has focused narrowly on Central-to-Northern dialect transfer using synthetic data and limited dialectal diversity. These efforts exclude Southern varieties and intra-regional variants within the North. We introduce ViDia2Std, the first manually annotated parallel corpus for dialect-to-standard Vietnamese translation covering all 63 provinces. Unlike prior datasets, ViDia2Std includes diverse dialects from Central, Southern, and non-standard Northern regions often absent from existing resources, making it the most dialectally inclusive corpus to date. The dataset consists of over 13,000 sentence pairs sourced from real-world Facebook comments and annotated by native speakers across all three dialect regions. To assess annotation consistency, we define a semantic mapping agreement metric that accounts for synonymous standard mappings across annotators. Based on this criterion, we report agreement rates of 86% (North), 82% (Central), and 85% (South). We benchmark several sequence-to-sequence models on ViDia2Std. mBART-large-50 achieves the best results (BLEU 0.8166, ROUGE-L 0.9384, METEOR 0.8925), while ViT5-base offers competitive performance with fewer parameters. ViDia2Std demonstrates that dialect normalization substantially improves downstream tasks, highlighting the need for dialect-aware resources in building robust Vietnamese NLP systems.},
  number={36},
  journal={Proceedings of the AAAI Conference on Artificial Intelligence},
  author={Anh Ta, Khoa and Van Dinh, Nguyen and Nguyen, Kiet Van},
  year={2026},
  month={Mar.},
  pages={29995-30004}
}

@inproceedings{srirag-etal-2025-evaluating,
    title = "Evaluating Dialect Robustness of Language Models via Conversation Understanding",
    author = "Srirag, Dipankar  and
      Sahoo, Nihar Ranjan  and
      Joshi, Aditya",
    booktitle = "Proceedings of the Second Workshop on Scaling Up Multilingual {\&} Multi-Cultural Evaluation",
    month = jan,
    year = "2025",
    address = "Abu Dhabi",
    publisher = "Association for Computational Linguistics",
    url = "https://aclanthology.org/2025.sumeval-2.3/",
    pages = "24--38"
}

@inproceedings{lin2025redial,
    title = "Assessing Dialect Fairness and Robustness of Large Language Models in Reasoning Tasks",
    author = "Lin, Fangru  and
      Mao, Shaoguang  and
      La Malfa, Emanuele  and
      Hofmann, Valentin  and
      de Wynter, Adrian  and
      Wang, Xun  and
      Chen, Si-Qing  and
      Wooldridge, Michael J.  and
      Pierrehumbert, Janet B.  and
      Wei, Furu",
    editor = "Che, Wanxiang  and
      Nabende, Joyce  and
      Shutova, Ekaterina  and
      Pilehvar, Mohammad Taher",
    booktitle = "Proceedings of the 63rd Annual Meeting of the Association for Computational Linguistics (Volume 1: Long Papers)",
    month = jul,
    year = "2025",
    address = "Vienna, Austria",
    publisher = "Association for Computational Linguistics",
    url = "https://aclanthology.org/2025.acl-long.317/",
    doi = "10.18653/v1/2025.acl-long.317",
    pages = "6317--6342",
    ISBN = "979-8-89176-251-0"
}

@inproceedings{gupta2025endive,
    title = "{E}n{D}ive: A Cross-Dialect Benchmark for Fairness and Performance in Large Language Models",
    author = "Gupta, Abhay  and
      Cheung, Jacob  and
      Meng, Philip  and
      Sayyed, Shayan  and
      Zhu, Kevin  and
      Liao, Austen  and
      O{'}Brien, Sean",
    editor = "Christodoulopoulos, Christos  and
      Chakraborty, Tanmoy  and
      Rose, Carolyn  and
      Peng, Violet",
    booktitle = "Findings of the Association for Computational Linguistics: EMNLP 2025",
    month = nov,
    year = "2025",
    address = "Suzhou, China",
    publisher = "Association for Computational Linguistics",
    url = "https://aclanthology.org/2025.findings-emnlp.913/",
    doi = "10.18653/v1/2025.findings-emnlp.913",
    pages = "16830--16855",
    ISBN = "979-8-89176-335-7"
}

@inproceedings{faisal2024dialectbench,
    title = "{DIALECTBENCH}: An {NLP} Benchmark for Dialects, Varieties, and Closely-Related Languages",
    author = "Faisal, Fahim  and
      Ahia, Orevaoghene  and
      Srivastava, Aarohi  and
      Ahuja, Kabir  and
      Chiang, David  and
      Tsvetkov, Yulia  and
      Anastasopoulos, Antonios",
    editor = "Ku, Lun-Wei  and
      Martins, Andre  and
      Srikumar, Vivek",
    booktitle = "Proceedings of the 62nd Annual Meeting of the Association for Computational Linguistics (Volume 1: Long Papers)",
    month = aug,
    year = "2024",
    address = "Bangkok, Thailand",
    publisher = "Association for Computational Linguistics",
    url = "https://aclanthology.org/2024.acl-long.777/",
    doi = "10.18653/v1/2024.acl-long.777",
    pages = "14412--14454"
}

@inproceedings{ziems2022multivalue,
  title={Multi-VALUE: A Framework for Cross-Dialectal English NLP},
  author={Ziems, Caleb and Held, William and Yang, Jingfeng and Dhamala, Jwala and Gupta, Rahul and Yang, Diyi},
  year={2022}
}

@inproceedings{altakrori2026dialectalarabicmmlu,
  title = {DialectalArabicMMLU: Benchmarking Dialectal Capabilities in Arabic and Multilingual Language Models},
  author = {Altakrori, Malik H. and Habash, Nizar and Lynn, Teresa and Samih, Younes and Freihat, Abed Alhakim and Chirkunov, Kirill and AbuOdeh, Muhammed and Florian, Radu and Nakov, Preslav and Aji, Alham Fikri},
  booktitle = {Proceedings of the Fifteenth Language Resources and Evaluation Conference (LREC 2026)},
  month = {May},
  year = {2026},
  pages = {3199--3219},
  address = {Palma, Mallorca, Spain},
  publisher = {European Language Resources Association (ELRA)},
  editor = {Piperidis, Stelios and Bel, Núria and van den Heuvel, Henk and Ide, Nancy and Krek, Simon and Toral, Antonio},
  doi = {10.63317/3cy68duew55b}
}

@inproceedings{mousi-etal-2025-aradice,
    title = "{A}ra{D}i{CE}: Benchmarks for Dialectal and Cultural Capabilities in {LLM}s",
    author = "Mousi, Basel  and
      Durrani, Nadir  and
      Ahmad, Fatema  and
      Hasan, Md. Arid  and
      Hasanain, Maram  and
      Kabbani, Tameem  and
      Dalvi, Fahim  and
      Chowdhury, Shammur Absar  and
      Alam, Firoj",
    editor = "Rambow, Owen  and
      Wanner, Leo  and
      Apidianaki, Marianna  and
      Al-Khalifa, Hend  and
      Eugenio, Barbara Di  and
      Schockaert, Steven",
    booktitle = "Proceedings of the 31st International Conference on Computational Linguistics",
    month = jan,
    year = "2025",
    address = "Abu Dhabi, UAE",
    publisher = "Association for Computational Linguistics",
    url = "https://aclanthology.org/2025.coling-main.283/",
    pages = "4186--4218"
}

@article{artstein-poesio-2008-survey,
    title = "Survey Article: Inter-Coder Agreement for Computational Linguistics",
    author = "Artstein, Ron  and
      Poesio, Massimo",
    journal = "Computational Linguistics",
    volume = "34",
    number = "4",
    year = "2008",
    url = "https://aclanthology.org/J08-4004/",
    doi = "10.1162/coli.07-034-R2",
    pages = "555--596"
}

@misc{qwen2025qwen25technicalreport,
      title={Qwen2.5 Technical Report}, 
      author={Qwen and : and An Yang and Baosong Yang and Beichen Zhang and Binyuan Hui and Bo Zheng and Bowen Yu and Chengyuan Li and Dayiheng Liu and Fei Huang and Haoran Wei and Huan Lin and Jian Yang and Jianhong Tu and Jianwei Zhang and Jianxin Yang and Jiaxi Yang and Jingren Zhou and Junyang Lin and Kai Dang and Keming Lu and Keqin Bao and Kexin Yang and Le Yu and Mei Li and Mingfeng Xue and Pei Zhang and Qin Zhu and Rui Men and Runji Lin and Tianhao Li and Tianyi Tang and Tingyu Xia and Xingzhang Ren and Xuancheng Ren and Yang Fan and Yang Su and Yichang Zhang and Yu Wan and Yuqiong Liu and Zeyu Cui and Zhenru Zhang and Zihan Qiu},
      year={2025},
      eprint={2412.15115},
      archivePrefix={arXiv},
      primaryClass={cs.CL},
      url={https://arxiv.org/abs/2412.15115}, 
}

@misc{grattafiori2024llama3herdmodels,
      title={The Llama 3 Herd of Models}, 
      author={Aaron Grattafiori and Abhimanyu Dubey and Abhinav Jauhri and Abhinav Pandey and Abhishek Kadian and Ahmad Al-Dahle and Aiesha Letman and Akhil Mathur and Alan Schelten and Alex Vaughan and Amy Yang and Angela Fan and Anirudh Goyal and Anthony Hartshorn and Aobo Yang and Archi Mitra and Archie Sravankumar},
      year={2024},
      eprint={2407.21783},
      archivePrefix={arXiv},
      primaryClass={cs.AI},
      url={https://arxiv.org/abs/2407.21783}, 
}

@misc{jiang2023mistral7b,
      title={Mistral 7B}, 
      author={Albert Q. Jiang and Alexandre Sablayrolles and Arthur Mensch and Chris Bamford and Devendra Singh Chaplot and Diego de las Casas and Florian Bressand and Gianna Lengyel and Guillaume Lample and Lucile Saulnier and Lélio Renard Lavaud and Marie-Anne Lachaux and Pierre Stock and Teven Le Scao and Thibaut Lavril and Thomas Wang and Timothée Lacroix and William El Sayed},
      year={2023},
      eprint={2310.06825},
      archivePrefix={arXiv},
      primaryClass={cs.CL},
      url={https://arxiv.org/abs/2310.06825}, 
}

@misc{gemmateam2024gemma2improvingopen,
      title={Gemma 2: Improving Open Language Models at a Practical Size}, 
      author={Gemma Team and Morgane Riviere and Shreya Pathak and Pier Giuseppe Sessa and Cassidy Hardin and Surya Bhupatiraju and Léonard Hussenot and Thomas Mesnard and Bobak Shahriari and Alexandre Ramé and Johan Ferret and Peter Liu and Pouya Tafti},
      year={2024},
      eprint={2408.00118},
      archivePrefix={arXiv},
      primaryClass={cs.CL},
      url={https://arxiv.org/abs/2408.00118}, 
}

@misc{gemmateam2025gemma3technicalreport,
      title={Gemma 3 Technical Report}, 
      author={Gemma Team and Aishwarya Kamath and Johan Ferret and Shreya Pathak and Nino Vieillard and Ramona Merhej and Sarah Perrin and Tatiana Matejovicova and Alexandre Ramé and Morgane Rivière},
      year={2025},
      eprint={2503.19786},
      archivePrefix={arXiv},
      primaryClass={cs.CL},
      url={https://arxiv.org/abs/2503.19786}, 
}

@article{chien2023vistral,
  author = {Chien Van Nguyen and
          Thuat Nguyen and
          Quan Nguyen and
          Huy Nguyen and
          Bj{\"o}rn Pl{\"u}ster and
          Nam Pham and
          Huu Nguyen and
          Patrick Schramowski and
          Thien Nguyen},
  title = {Vistral-7B-Chat - Towards a State-of-the-Art Large Language Model for Vietnamese},
  year = 2023,
}

@inproceedings{zhang-etal-2025-seallms,
    title = "{S}ea{LLM}s 3: Open Foundation and Chat Multilingual Large Language Models for {S}outheast {A}sian Languages",
    author = "Zhang, Wenxuan  and
      Chan, Hou Pong  and
      Zhao, Yiran  and
      Aljunied, Mahani  and
      Wang, Jianyu  and
      Liu, Chaoqun  and
      Deng, Yue  and
      Hu, Zhiqiang  and
      Xu, Weiwen  and
      Chia, Yew Ken  and
      Li, Xin  and
      Bing, Lidong",
    editor = "Dziri, Nouha  and
      Ren, Sean (Xiang)  and
      Diao, Shizhe",
    booktitle = "Proceedings of the 2025 Conference of the Nations of the Americas Chapter of the Association for Computational Linguistics: Human Language Technologies (System Demonstrations)",
    month = apr,
    year = "2025",
    address = "Albuquerque, New Mexico",
    publisher = "Association for Computational Linguistics",
    url = "https://aclanthology.org/2025.naacl-demo.10/",
    doi = "10.18653/v1/2025.naacl-demo.10",
    pages = "96--105",
    ISBN = "979-8-89176-191-9"
}
\end{document}